\documentclass{article}

\usepackage{arxiv}

\usepackage[utf8]{inputenc} 
\usepackage[T1]{fontenc}    
\usepackage{hyperref}       
\usepackage{url}            
\usepackage{booktabs}       
\usepackage{amsfonts}       
\usepackage{nicefrac}       
\usepackage{microtype}      
\usepackage{lipsum}

\usepackage{times}
\usepackage{epsfig}
\usepackage{graphicx}
\usepackage{amsmath}
\usepackage{amssymb}

\usepackage{xcolor}

\title{Ear2Face: Deep Biometric Modality Mapping}

\author{Dogucan Yaman\textsuperscript{*} \\ Karlsruhe Institute of Technology \\ \texttt{dogucan.yaman@kit.edu}
  \And
  Fevziye Irem Eyiokur\textsuperscript{*} \\
  Karlsruhe Institute of Technology \\
  \texttt{fevziye.yaman@kit.edu} 
  \And
  Haz{\i}m Kemal Ekenel \\
  Istanbul Technical University \\
  \texttt{ekenel@itu.edu.tr} 
}

\begin{document}

\maketitle

\thispagestyle{empty}
\renewcommand{\thefootnote}{\Alph{footnote}}
\footnotetext{\textsuperscript{*}The authors have equally contributed and conducted this study while they were students at Istanbul Technical University.}

\begin{abstract}
In this paper, we explore the correlation between different visual biometric modalities. For this purpose, we present an end-to-end deep neural network model that learns a mapping between the biometric modalities. Namely, our goal is to generate a frontal face image of a subject given his/her ear image as the input. We formulated the problem as a paired image-to-image translation task and collected datasets of ear and face image pairs from the Multi-PIE and FERET datasets to train our GAN-based models. We employed feature reconstruction and style reconstruction losses in addition to adversarial and pixel losses. 
We evaluated the proposed method both in terms of reconstruction quality and in terms of person identification accuracy. To assess the generalization capability of the learned mapping models, we also run cross-dataset experiments. That is, we trained the model on the FERET dataset and tested it on the Multi-PIE dataset and vice versa. We have achieved very promising results, especially on the FERET dataset, generating visually appealing face images from ear image inputs. Moreover, we attained a very high cross-modality person identification performance, for example, reaching 90.9\% Rank-10 identification accuracy on the FERET dataset.
\end{abstract}


\section{Introduction}
A desirable technology for biometric identification would be to be able to match different biometric modalities to each other. For example, recent speech2face \cite{oh2019speech2face} study proposed a method to generate face images from speech input. This way one can compare speech signals with the face images in the gallery, performing cross-modal person identification. Although audio-visual person identification datasets are abundant, since speech and image signals have different characteristics, it is still very challenging to learn a mapping between them. On the other hand, there are various visual biometric modalities, such as fingerprint, ear, iris, hand, and face, and since they are all from the visual domain, it would be easier to learn a mapping between them. In this paper, we focus on learning a deep mapping between face and ear modalities. We opted for using face and ear modalities, since multimodal person identification datasets are rather limited and since we were able to collect a large amount of ear-face image pairs by utilizing the Multi-PIE \cite{multi_pie} and FERET \cite{FERET_dataset} datasets.



In human nature, genotype is one of the main factors that determines the face \cite{peng2013detecting, claes2014modeling, richmond2018facial, srinivas2017dna2face} and other biometric parts. 
Several works have shown that the relationship between DNA information and face appearance can be established \cite{peng2013detecting, claes2014modeling, srinivas2017dna2face, richmond2018facial, crouch2018genetics, sero2019facial}.
Since both ear and face are biometric traits of an individual and their phenotypes are generated based on genotypes, we expect an implicit relationship between ear and face through genetic knowledge \cite{srinivas2017dna2face, crouch2018genetics, richmond2018facial, sero2019facial}. These previous works have motivated us to investigate the correlation between different visual biometric modalities, namely, ear and face, and learn a mapping between them in order to have a cross-modal biometric identification system.


\begin{figure}[t!]
\begin{center}
\includegraphics[scale=0.59]{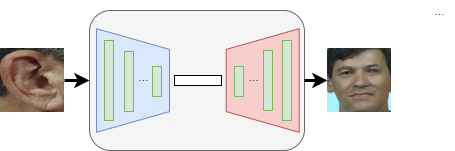}
\end{center}
\caption{The generative \textit{Ear2Face} network learns a mapping between ear and face images.}
\label{fig:task}
\end{figure}

In this work, we focus on learning this relationship between different modalities, ear and face, using generative adversarial networks (GAN). After generative adversarial network was proposed, it has demonstrated superior performance on many different tasks, such as image to image translation, e.g., sketch2image \cite{hu2017now,sangkloy2017scribbler,chen2018sketchygan}, style transfer \cite{huang2017arbitrary,chen2018gated}, and cross-modal learning \cite{kim2017learning}, e.g., speech2face \cite{oh2019speech2face}, DNA2Face \cite{srinivas2017dna2face}. The most relevant works to our study are \cite{oh2019speech2face,wav2pix2019icassp}, in which authors aim at learning the mapping between voice and face modalities. 

The proposed GAN model, named as \textit{Ear2Face}, takes an ear image as input and generates a frontal face image as shown in Figure~\ref{fig:task}. We formulated the problem as an image-to-image translation task and developed our model based on the model proposed in \cite{isola2017image}. Besides, we benefited from feature and style reconstruction losses to enhance the learning capacity of the network. 
We summed up adversarial loss, pixel loss, feature reconstruction loss, and style reconstruction loss in the objective function and we presented $\beta$ and $\gamma$ coefficients to restrict the effect of feature and style reconstruction losses on overall loss function. We also employed $\lambda$ coefficient for pixel loss as in \cite{isola2017image}. For the experiments, we collected ear and frontal face image pairs from the Multi-PIE \cite{multi_pie} and the FERET \cite{FERET_dataset} datasets, which are popular face datasets that also contain profile views of the subjects. 
Afterwards, since we need to preserve identity information during reconstruction, we matched ear and face images of each subject to perform paired training. Later, we fed the network with these image pairs. In the training phase, we utilized ResNet-50 \cite{he2016deep} deep CNN model with weights of VGGFace2 \cite{cao2018vggface2}, which is a robust face recognition model, to extract features from reconstructed and ground truth images to measure feature reconstruction and style reconstruction losses. In the end, we evaluated our model capability with five different metrics in terms of quality of generated face images. We also conducted face recognition experiments using the reconstructed face images. 

Our contributions can be summarized as follows:
\begin{itemize}
    \item We presented a novel study on biometric modality mapping. We formulated the problem as an image-to-image translation between images of different biometric modalities. 
    \item We showed that the genotype-based implicit relationship between ear and face can be learnt via GANs.
    \item We created a GAN model based on \cite{isola2017image} and added face reconstruction and style reconstruction losses, in addition to adversarial loss and pixel loss to improve the quality of generated images. 
    \item We used five different metrics to evaluate our method's reconstruction performance. Besides, we conducted face recognition experiments using the reconstructed face images. Moreover, we performed cross-dataset experiments to assess the generalization capacity of our deep mapping model.
    \item We showed that our model generates not only perceptually appealing face images but also preserves identity information.
\end{itemize}

    The rest of the paper is organized as follows: In Section~2, we give brief information about previous work on ear biometrics, image-to-image translation, and  cross-modal learning.  In Section 3, we explain the proposed network, used loss functions, and the training procedure. In Section 4,  we first present the datasets and evaluation metrics,  then, we discuss the results of  Ear2Face  model  on  both  datasets. We also present and discuss the obtained face  recognition  results  using the  reconstructed  face  images.  Finally, we conclude our work in Section 5.

\section{Related Work}
\textbf{Ear biometrics.} 
Ear images have been utilized in many different works for the purpose of person identification \cite{more_than_ziga,emervsivc2017training,our_ear_journal}, age estimation \cite{yaman2018age,yaman2019multimodal}, and gender classification \cite{pflug2012ear,abaza2013survey,gnanasivam2013gender,khorsandi2013gender,lei2013gender,yaman2018age,yaman2019multimodal}. All these works show the usefulness and effectiveness of ear as a biometric trait. Besides, there exists some works \cite{zhang2011hierarchical,yaman2019multimodal} that utilize both profile face and ear images together to improve the biometric system's performance. The results indicate that profile face and ear have complementary features and using them together leads to a performance improvement, especially for the age and gender predictions.

\textbf{Image2image translation.} After generative adversarial network \cite{goodfellow2014generative} is proposed, it has been adapted for different tasks beyond artificial image generation from noise. Particularly, image to image translation is one of the important and popular fields in generative works. GANs are used in many different image to image translation works, such as domain transfer \cite{isola2017image,zhu2017unpaired,choi2018stargan,zhang2018self,tang2019multi}, super resolution \cite{ledig2017photo,lim2017enhanced,wang2018high}, style transfer \cite{huang2017arbitrary,chen2018gated}, and etc. All these studies show the effectiveness of the GAN models in terms of learning a mapping between different domains for different purposes.

\textbf{Cross-modal learning.} The high performance of generative models on image generation and image to image translation tasks led its use in other areas, such as cross-modal learning. Image generation from text \cite{reed2016generative,dash2017tac,zhang2017stackgan,xu2018attngan,zhu2019dm} and audio \cite{chen2017deep,qiu2018image,hao2018cmcgan,wav2pix2019icassp,duan2019cascade,oh2019speech2face,wan2019towards,deng2020unsupervised} are the most common examples of cross-modal learning. The idea is to explore the relationship in the feature space between different modalities with generative models to translate these modalities to each other. In \cite{wav2pix2019icassp}, face images are reconstructed from speech data using conditional GANs in an end-to-end manner. In \cite{oh2019speech2face}, authors aim at learning mapping between face features and speech features. They converted speech data to the spectrogram format to feed voice encoder in order to embed it into feature space. In the system, face and voice data are used as paired and features are extracted from both spectrogram and face image. While encoder network that is employed for embedding voice data is a trainable part of the system, the other parts, feature extractor and decoder for the face reconstruction models are well-known pretrained models. Moreover, features of voice and face data are utilized to calculate loss during training. In \cite{qiu2018image} music data is employed to generate scene images which represent what related audio makes feel. In \cite{wan2019towards}, GAN-based model is employed to create more qualified audio-image pairs. Other studies \cite{chen2017deep,hao2018cmcgan,duan2019cascade} focus on audio data, which belongs to musical instruments, to reconstruct a scene with related instrument using GANs. Two different GAN models are developed to generate images from audio and audio from image in \cite{chen2017deep}. Unlike the previous study, in \cite{hao2018cmcgan}, a combined cyclic generative adversarial network, which is named as CMCGAN, is proposed. Lastly, to enhance the quality of the coarse outputs and obtain fine-grained results, authors provided two-stage GAN network in \cite{duan2019cascade}. For a detailed review of audio-image translation tasks, please refer to a recent survey \cite{zhu2020deep}. 

\section{Face Reconstruction}
\begin{figure*}[b!]
\begin{center}
\includegraphics[scale=0.4]{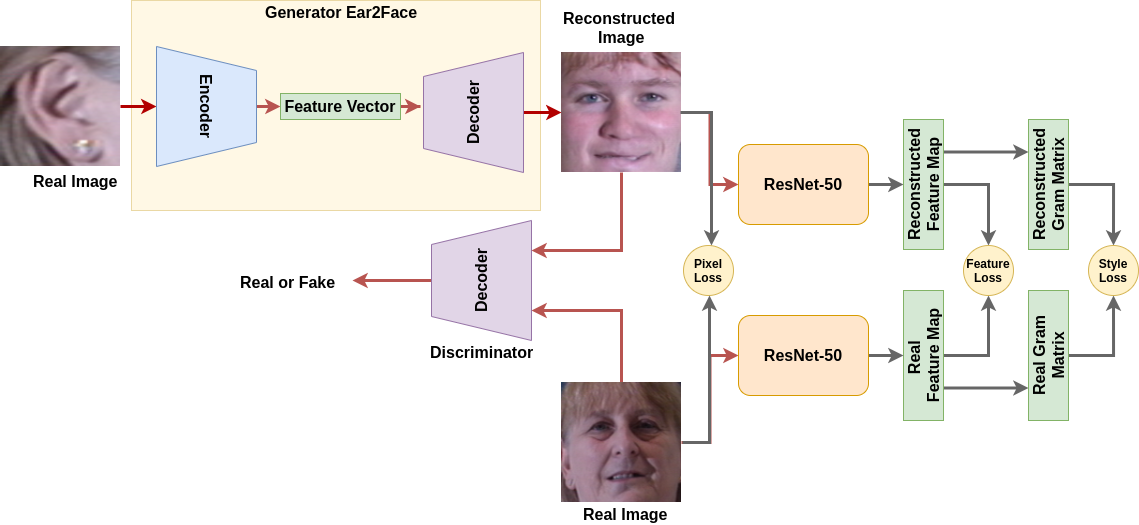}
\end{center}
\caption{Proposed \textit{Ear2Face} network. The network takes an ear image as input and learns face reconstruction from it. Adversarial loss, pixel loss, face reconstruction loss \cite{johnson2016perceptual}, and style reconstruction loss \cite{gatys2015texture, gatys2015neural} are employed together in training. For the feature and style reconstruction losses, features are extracted with ResNet-50 architecture \cite{he2016deep} using VGGFace2 pretrained model \cite{cao2018vggface2}. The extracted features are obtained from global average pooling layer and its dimension is $[1 \times 2048]$. }
\label{fig:proposed_network}
\end{figure*}

In this section, we explain the proposed GAN model and employed loss functions in the training. The proposed \textit{Ear2Face} network is shown in Figure \ref{fig:proposed_network}. While generator part of this model takes an ear image as an input and tries to reconstruct face data, discriminator is trained with real images and is responsible to discriminate between real and fake data. Pretrained VGGFace2 \cite{cao2018vggface2} is employed for feature extraction, and pixel loss, feature reconstruction loss \cite{johnson2016perceptual}, and style reconstruction loss \cite{gatys2015texture, gatys2015neural} are measured between generated face image and real face image in both pixel space and feature space.

\subsection{Model}
For our deep biometric modality mapping system, we employed a GAN architecture based on pix2pix model \cite{isola2017image}, and named our model as \textit{Ear2Face}. In \cite{isola2017image}, generator network is adjusted based on U-Net architecture \cite{ronneberger2015u} and skip connections are included. For the discriminator, similar architecture with \cite{li2016precomputed} is employed. 


While conditional generator network struggles to generate artificial data that can deceive the discriminator, discriminator network tries to learn training data distribution and it is responsible to discriminate between real and fake data. In this work, the generator fetches random noise and source image as an input and then it learns the relationship between target image and input data, $G: \{x,z\} \rightarrow y $. 

\subsection{Loss functions}
\textbf{Adversarial loss.} The objective function of the conditional GAN is

\begin{equation}
\label{eq_adversarial_loss}
    L_{cGAN}(G, D) = \mathop{\mathbb{E}}_{x,y}[log(D(x,y)] + \mathop{\mathbb{E}}_{x,z}[log(1-D(x,G(x,z))]
\end{equation}
where G is a generator and D is a discriminator network. While unconditional GAN tries to generate artificial data from random noise, z, conditional GAN gets additional input which is a source image in this work and represented as \textit{x} in Equation \ref{eq_adversarial_loss}. 

\textbf{Pixel loss.} In addition to previous cost function, we also included an additional function, which is L1 distance as in~\cite{isola2017image} and expressed in Equation \ref{eq_l1_loss}. This function is responsible to compare generated data and real data in the pixel space and thus, it forces the network to generate analogous samples with target data. Since several papers revealed that using L2 distance caused blurry images, L1 distance is employed. 

\begin{equation}
\label{eq_l1_loss}
    L_{L1}(G) = \mathop{\mathbb{E}}_{x,y,z}[||y-G(x,z)||_1]
\end{equation}

In this equation, while \textit{y} represents target (real) image, \textit{G(x,z)} is the reconstructed face image.

\textbf{Feature reconstruction loss.} In order to compare generated and real images in a feature space, feature reconstruction loss is added to the objective function. The intention of using feature reconstruction loss is to stimulate network to learn similar feature representation with target data in order to retain structure and content \cite{johnson2016perceptual}. The feature reconstruction loss is defined as

\begin{equation}
\label{eq_feature_loss}
    L_{feat}(G(x,z)) =  \mathop{\mathbb{E}}_{x,y,z} \Big[ \frac{1}{C_jH_jW_j}||\o_j(G(x,z)) - \o_j(y)||_2^2 \Big]
\end{equation}
where $C_j, H_j,$ and $W_j$ are number of channels, width, and height of the image, respectively. While $\psi$ represents the model that is employed for feature extraction, $j$ is the layer of $\psi$ that features are obtained from. The normalized root of the Euclidean distance between features of generated image and real image are calculated as a feature reconstruction loss. In our system, pretrained VGGFace2 model~\cite{cao2018vggface2} is employed, which is based on ResNet-50 architecture \cite{he2016deep}. Features are extracted from global average pooling layer of this model. The dimension of feature vector is $[1 \times 2048]$.

\textbf{Style reconstruction loss}.
In addition to the feature reconstruction loss, style reconstruction loss \cite{gatys2015texture, gatys2015neural} is utilized as well. The main motivation behind employing this loss function is to charge distinctness between fake and real images in terms of their style, such as textures and colors. In order to compute style reconstruction loss, Gram matrix is needed to be calculated beforehand. The Gram matrix formulation is

\begin{equation}
\label{eq_gram_matrix}
    Gram(\psi) = \frac{\psi \psi^T}{C_jH_jW_j}
\end{equation}
where $\psi$ is a feature map that is acquired from global average pooling layer of the ResNet-50 model \cite{he2016deep} as in feature reconstruction loss. The feature map and its transpose are multiplied and then normalized to obtain Gram matrix. This is repeated with generated image feature map and target image feature map. Afterwards, style reconstruction loss is calculated using these Gram matrices as follows:

\begin{equation}
\label{eq_style_loss}
    L_{style}(G(x,z), y) = \mathop{\mathbb{E}}_{x,y,z} ||Gram_j^{\o}(G(x,z)) - Gram_j^{\o}(y)||_{F}^2
\end{equation}

In this equation, $j$ is the layer of ResNet-50 model ($\psi$) for feature extraction as in feature reconstruction loss. Afterward, the extracted feature maps are forwarded to perform Gram matrices calculation using Equation \ref{eq_gram_matrix}. The Gram matrix is calculated both for fake and real images, thereafter the style reconstruction loss is handled via computing squared Frobenius norm (F-norm) of the outcome of subtraction of Gram matrices.

\begin{figure*}[t!]
\begin{center}
\includegraphics[scale=0.45]{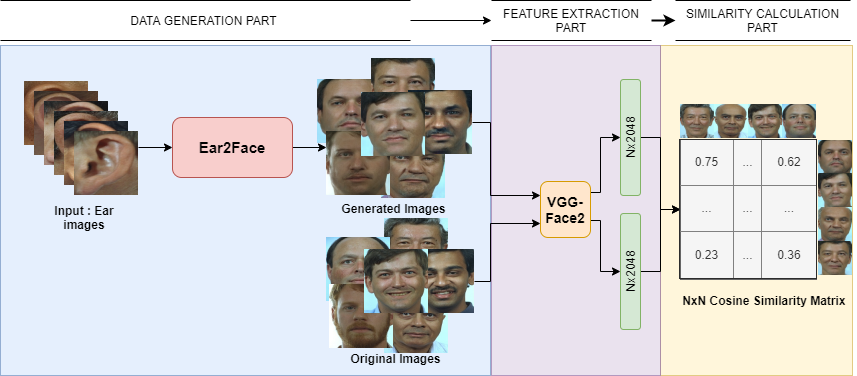}
\end{center}
\caption{Face recognition scheme. Firstly, we reconstructed frontal face images from ear data using the proposed Ear2Face model. Then, the ResNet-50 model with VGG-Face2 weights are employed to extract features from reconstructed and original face images. Features are obtained from global average pooling layer and of size $1 \times 2048$. Later, cosine similarity between these feature vectors are calculated and a similarity matrix is constructed based on these values. Finally, face recognition performance is measured by processing the similarity matrix.}
\label{fig:test_procedure}
\end{figure*}

\subsection{Training procedure}
The final objective function of the proposed method is presented in Equation \ref{eq_final_cost_function}.

\begin{equation}
\label{eq_final_cost_function}
    G^* = arg \min_{G} \max_{D} L_{cGAN}(G,D) + \lambda L_{L1}(G)
    + \beta L_{feat}(G) + \gamma L_{style}(G)
\end{equation}

This formula is the combination of adversarial loss, L1 loss, feature reconstruction loss, and style reconstruction loss. All these loss terms, except adversarial loss, are multiplied with the corresponding coefficients, $\lambda, \beta, \gamma$. 
According to empirical evaluation, we set the $\lambda, \beta,$ and $\gamma$ parameters to 10, 0.25, 0.1, respectively. Finally, the optimization objective of this function is to minimize generator, G, and maximize discriminator, D.

\subsection{Face recognition}
In order to assess whether the reconstructed face images are able to preserve identity information, hence useful for person identification, we have conducted face recognition experiments.
The proposed face recognition scheme is shown in Figure \ref{fig:test_procedure}. In the experiments, we employed pretrained ResNet-50 CNN model that was trained on VGGFace2 dataset~\cite{cao2018vggface2} for feature extraction. We extracted features both from reconstructed face images, which correspond to the probe set, and original face images, which correspond to the gallery set. Then, the cosine similarity between 
the feature vectors of each image from gallery and probe sets are calculated to generate $N \times N$ similarity matrix. Afterwards, the face recognition accuracy is calculated using this similarity matrix.

\section{Experimental Results}
In this section, we presented the datasets and evaluation metrics that we used in our experiments. We, then, provided and discussed face reconstruction and face recognition performance. Finally, we shared cross-dataset experiment results to quantify proposed model's generalization capability. 

\subsection{Datasets}
In this work, we generated paired images both from the Multi-PIE \cite{multi_pie} and the FERET \cite{FERET_dataset} datasets. 
We followed a similar approach to compose ear-face image pairs from these datasets. We first executed OpenCV \cite{opencv} ear detection algorithm and dlib face detector \cite{dlib} to capture ear and frontal face images. Afterwards, we resized the detected ear and face images to the same size, which is $256 \times 256$.

\subsubsection{Multi-PIE}

In Multi-PIE dataset \cite{multi_pie}, we detected and created 8533 ear-frontal face image pairs belonging to 250 subjects. We separated Multi-PIE dataset into training and three different test sets. For the training set, we selected 240 out of 250 subjects and obtained 6544 ear-face image pairs. Remaining 10 subjects are utilized for subject independent test set. Overall, we have two subject dependent and one subject independent test sets. The details about them are explained below.

\textbf{Subject independent (S.ID.) test set.} We used remaining 10 subjects, who are not in the training set. In the training, the proposed model did not see and learn these 10 subjects and this way, we investigated the subject independent performance of the model.

\textbf{Subject dependent (S.D.) test set 1.} In this set, there are 1677 images of 240 subjects. These 240 subjects are the same with the subjects in the training set, however, the images are different. That is, there are no common images in the training and test sets.

\textbf{Subject dependent (S.D.) test set 2.} 10 subjects are randomly selected from the subject dependent test set 1. The purpose of this set is to create a subject dependent test set that has the same number of subjects with the subject independent test set, in order to compare subject dependent and subject independent results fairly.

\begin{table}[h!]
\begin{center}
\begin{tabular}{c|cc|cc|c}
\hline
Dataset & \multicolumn{2}{c}{Training set} & \multicolumn{2}{c}{Test set} & Test set name \\
\hline
 & \# of sub. & \# of img. & \# of sub. & \# of img. & \\ 
\hline
Multi-PIE & 240 & 6544 & 240 & 1677 & Subject dependent test set 1\\
Multi-PIE & 240 & 6544 & 10 & 95 & Subject dependent test set 2 \\
Multi-PIE & 240 & 6544 & 10 & 312 & Subject independent \\
FERET & 504 & 623 & 504 & 504 & Subject dependent test set 1\\
FERET & 504 & 623 & 55 & 55 & Subject dependent test set 2 \\
FERET & 504 & 623 & 55 & 55 & Subject independent \\
\hline
\end{tabular}
\end{center}
\caption{The training and test sets generated from the Multi-PIE and the FERET datasets. The training set contains 80\% of the images from 240 training subjects for the Multi-PIE dataset and 80\% of the images from 504 training subjects for the FERET dataset. The remaining subjects for both datasets are selected to generate subject independent test sets. For subject dependent test set 1, we used remaining 20\% of the images from 240 subjects for the Multi-PIE and 20\% of the images from 504 subjects for the FERET dataset. In order to have a fair comparison between the subject dependent and subject independent setups, we also randomly selected the same number of subjects with the subject independent test set to create subject dependent test set 2. The experimental setup and codes will be available \textcolor{magenta}{\url{https://github.com/yamand16/ear2face}}. }
\label{table:test_set}
\end{table}

\subsubsection{FERET}

For the FERET dataset, we obtained 1182 ear-face image pairs from 559 different subjects. 
In this dataset, while 504 subjects have more than one ear-face image pair, the remaining 55 subjects have only one ear-face image pair. One image from 504 subjects are selected for the subject dependent test set and remaining images are used in the training set.

\textbf{Subject independent (S.ID.) test set.} As mentioned above, 55 subjects have only one ear-face pair. Because of that, we selected these 55 subjects for the subject independent test.

\textbf{Subject dependent (S.D.) test set 1.} 504 images belonging to 504 subjects are chosen for this set. Each subject has one ear-face image pair, which is not included in the training set.

\textbf{Subject dependent (S.D.) test set 2.} In order to have the same number of subjects with the subject independent test set, 
we created this subset by randomly selecting 55 subjects from the subject dependent test set 1. This way, we obtain 55 ear-face image pairs belonging to 55 subjects that are not in the training set. 



We summarize the related information about the training and test sets of both datasets in Table \ref{table:test_set}. In addition to these experiments, we also performed cross-dataset experiments to explore the generalization capacity of the learned models. That is, we tested a model, that was trained on the training set of Multi-PIE dataset, on the test sets of the FERET dataset, and vice versa. Moreover, using the same setups in Table \ref{table:test_set}, we applied face recognition experiments using generated face images and original face images to investigate whether the identity information is preserved during reconstruction.



\subsection{Evaluation metrics}

\begin{table*}[t!]
\begin{center}
\begin{tabular}{ccc|ccccc}
\hline
Dataset & Test set & \# of subject & Pixel difference & Feature difference & Style difference & PSNR & SSIM \\
\hline
Multi-PIE & S.D. & 240 & 0.18 & 2.30 & 2.64 & 28.88 & 0.57 \\
Multi-PIE & S.ID. & 10 & 0.20 & 2.36  & 2.99 & 28.60 & 0.55 \\
\hline
FERET & S.D. & 504 & 0.16 & 1.86 & 1.81 & 28.53 & 0.61 \\
FERET & S.ID. & 55 & 0.17 & 1.88 & 1.82 & 28.44 & 0.60 \\
\hline
\end{tabular}
\end{center}
\caption{Evaluation results on Multi-PIE \cite{multi_pie} and FERET \cite{FERET_dataset} datasets. While the first column contains dataset, second column shows used test set which are subject independent set (S.ID.) and subject dependent set (S.D.). Third column shows the number of subject in the corresponding test set and the following columns indicate evaluation results with five different metrics.}
\label{table:all_results}
\end{table*}

\begin{figure*}[t!]
\begin{center}
\includegraphics[scale=0.30]{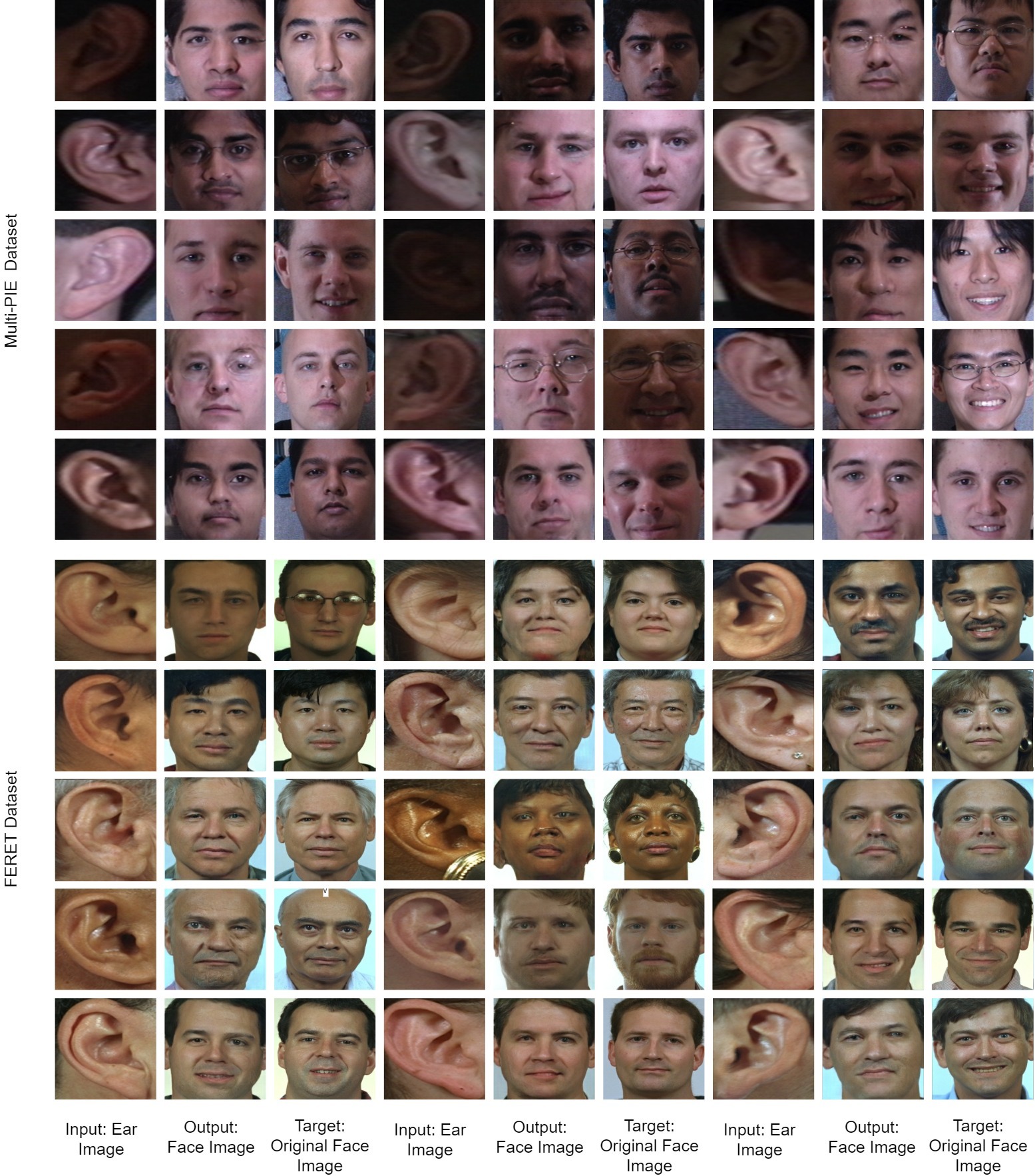}
\end{center}
\caption{Sample reconstructed images from the Multi-PIE \cite{multi_pie} (first five rows) and FERET \cite{FERET_dataset} (last five rows) datasets. In the first 2 groups which means first 6 columns are from subject dependent test set 1 and remaining 3 columns are from subject independent test set for both datasets. While output images are reconstructed face images using input ear images, target images represent ground truth face images. All these examples are obtained from test set. Although there are local corruptions over faces like eye location shifts, artifacts from accessories, the general qualitative results are satisfactory.}
\label{fig:samples}
\end{figure*}

In order to quantitatively assess the reconstruction performance, we employed five evaluation metrics, which are pixel similarity, feature similarity, style similarity, Peak Signal-to-Noise Ratio (PSNR), and Structural Similarity Index (SSIM) between the reconstructed and ground truth image. 

\textbf{Pixel difference.} We calculated L1 distance between ground truth and reconstructed face image to measure the similarity between them in the pixel space.

\textbf{Feature difference.} Besides comparing ground truth and reconstructed face images in the pixel space, we also benefited from feature similarity to compare them in the feature space. We employed Equation \ref{eq_feature_loss} to measure squared Euclidean distance between features of reconstructed and real face images.

\textbf{Style difference.} We also used features of reconstructed and real face images in order to calculate style differences between them. We used Equation \ref{eq_style_loss} to calculate the style difference.


\textbf{PSNR and SSIM.} 
While PSNR measures the numerical similarity between images via calculating the ratio between a range of the pixel value and Euclidean distance between pixels of the generated and real images, SSIM checks structural similarity between them.  

\subsection{Reconstruction Results}

Results of the subject dependent and subject independent experiments are presented in Table \ref{table:all_results}. While the first column contains the used dataset, second column indicates whether the test is subject dependent (S.D.) or subject independent (S.ID.). Following column shows the number of subjects in the test set and other columns contain evaluation scores. 
Subject dependent test set contains the same subjects with the training set but different images of them. On the other hand, subject independent test set includes different subjects than the ones in the training set.

\textbf{Quantitative evaluation.}
According to the experimental results in Table \ref{table:all_results}, results on the Multi-PIE subject dependent and subject independent experiments are similar in all five evaluation metrics. This outcome indicates that our model performs similarly on the subjects, who do not exist in the training set. Especially, relatively high PSNR and SSIM indicates the high face reconstruction capability of the network from input ear images.

We performed the same experiments on the FERET dataset. As in the Multi-PIE experiments, the subject dependent results are slightly better than subject independent ones in terms of considered evaluation metrics. Besides, both subject dependent and subject independent results with all metrics, except PSNR, are better than the Multi-PIE results. Especially, feature and style differences are significantly lower than the ones obtained on the Multi-PIE dataset. One reason for this could be the higher number of subjects available in the FERET dataset, which might have led to a better modelling of the appearance variations.

In Table \ref{table:all_results}, the subject dependent results represent both the \textit{subject dependent test set 1} and the \textit{subject dependent test set 2}. Since the results are almost the same for this experiment, we did not present the \textit{subject dependent test set 2} in a different row.


\textbf{Qualitative evaluation.}
Example images from the Multi-PIE and FERET datasets are presented in Figure \ref{fig:samples}. In these figures, input represents used ear image for the face reconstruction, output is the reconstructed face image from input ear image, and target contains real face image of the corresponding subject. These example images are from the test set. While first 6 columns are from subject dependent test set 1, other 3 columns are selected from subject independent test set. When reconstructed face images are examined, it seems that they have some partial deterioration, such as asymmetric face, eyes, partial skewness in the eye, nose, or mouth. Despite such slight skewness or asymmetric face, reconstructed face images resemble the original face images and there are very few artifacts on the image. Besides, when it is considered that all of these face images are reconstructed from only ears using relatively a small dataset, they are very promising. 
Moreover, texture quality of the generated face images, especially for the FERET dataset, are satisfactory in terms of similarity with the ground truth images and real human skin.


\begin{table*}[t!]
\begin{center}
\begin{tabular}{cccc|cccccc}
\hline 
Dataset & Images & Test set & \# Subj. & Rank-1 & Rank-2 & Rank-5 & Rank-10 & Rank-20 \\
\hline
Multi-PIE & Reconstructed & S.D. & 240 & 3\%  & 4.5\% & 6.6\% & 10\% & 14.3\% \\
Multi-PIE & Reconstructed & S.D. & 10 & 36.2\% & 40.4\% & 51.1\% & 60.6\% & 73.4\% \\
Multi-PIE & Reconstructed & S.ID. & 10 & 26.4\% & 30.5\% & 37.9\% & 43.1\% & 53.7\% \\
Multi-PIE & Real & S.D. & 240 & 99.1\% & 99.1\% & 99.1\% & 100\% & 100\% \\ 
\hline
FERET & Reconstructed & S.D. & 504 & 21.4\% & 27.1\% & 37.6\% & 47.3\% & 60.3\% \\
FERET & Reconstructed & S.D. & 55 & 64.5\% & 72.6\% & 83.8\% & 88.7\% & 94.3\% \\
FERET & Reconstructed & S.ID. & 55 & 47.3\% & 60\% & 85.5\% & 90.9\% & 98.2\% \\
FERET & Real & S.D. & 504 & 83.8\% & 83.8\% & 83.8\% & 100\% & 100\% \\
\hline
\end{tabular}
\end{center}
\caption{Face recognition results on the Multi-PIE \cite{multi_pie} and FERET \cite{FERET_dataset} datasets. The first column contains the dataset name, while the second column includes the type of test images. In the second column, reconstructed means reconstructed face images from ear, real means original face images. Moreover, next column, test set, represents which test set is employed for the corresponding experiments, e.g., subject dependent (S.D) or subject independent (S.ID). Next column is the number of subject and following five columns show face recognition accuracies.}
\label{table:face_recognition}
\end{table*}

\subsection{Face Recognition Results}

We present face recognition results on both datasets in Table \ref{table:face_recognition}. While images column indicates the employed images for the recognition task, the next column shows test set. Again, S.D. and S.ID. abbreviations represent subject dependent and subject independent sets, respectively. We also performed subject dependent experiment on a small subset of subject dependent test set 1 to have the same number of subjects with the S.ID. set in order to have a fair comparison with subject independent test results. 

According to the experimental results, face recognition accuracies on the real data ---frontal face images from the datasets--- are extremely high on both datasets, especially on the Multi-PIE dataset. On the other hand, face recognition performance using reconstructed face images is also very promising. Since, we obtained better reconstruction performance on the FERET dataset, this also led to better face recognition accuracies on the FERET dataset. For example, we reached 88.7\% and 90.9\% Rank-10 identification accuracies on the FERET dataset subject dependent and independent setups, respectively. Since the FERET dataset contains more subjects, it can cover more identity variations. On account of this, both in face reconstruction and face recognition experiments the subject dependent and independent setup results have been very close to each other. On the other hand, we attained 60.6\% and 43.1\% Rank-10 identification accuracies on the Multi-PIE subject dependent and independent setups, respectively. The larger performance gap between subject dependent and independent experiment results could be due to the fact that Multi-PIE dataset contains less number of subjects in the training, therefore, having a limited capacity to model identity variations.



\subsection{Cross-dataset Results}
In order to measure the generalization performance of the proposed network, we conducted cross-dataset experiments and presented their results in this section. In Table \ref{table:cross_test_reconstructed}, the first column, dataset, indicates the test dataset while model column states the dataset that was used in the model training phase. Subject dependent and subject independent represent used test sets of the corresponding dataset. As in Table \ref{table:all_results}, we did not mention test results on the \textit{subject dependent test set 2} in Table \ref{table:cross_test_reconstructed}, since the results are almost the same with the ones on the \textit{subject dependent test set 1}.

\begin{table*}[t!]
\begin{center}
\begin{tabular}{ccc|ccccc}
\hline
Dataset & Model & Test set & Pixel difference & Feature difference & Style difference & PSNR & SSIM \\
\hline
Multi-PIE & FERET & S.D. & 0.19 & 2.24 & 3.64 & 28.45 & 0.57 \\
Multi-PIE & FERET & S.ID. & 0.21 & 2.31 & 4.67 & 28.29 & 0.56 \\
\hline
FERET & Multi-PIE & S.D. & 0.21 & 2.34 & 6.21 & 27.89 & 0.54 \\
FERET & Multi-PIE & S.ID. & 0.21 & 2.44 & 6.92 & 28.03 & 0.56 \\
\hline
\end{tabular}
\end{center}
\caption{Cross-dataset evaluation results on the Multi-PIE \cite{multi_pie} and FERET \cite{FERET_dataset} datasets. While the first column contains dataset, the second one shows the used model for reconstructing images from corresponding dataset. Third one represents test setup and following columns indicate evaluation results with five different metrics.}
\label{table:cross_test_reconstructed}
\end{table*}


According to the cross-dataset results on the Multi-PIE and FERET datasets in Table \ref{table:cross_test_reconstructed}, it is clearly shown that the proposed model performs also well on the cross-dataset setup. Except style difference metric, all other metrics showed a similar performance with the corresponding dataset experiments. On the other hand, style difference metric gave poor results, 6.21 and 6.92 for the FERET, 3.64 and 4.67 for the Multi-PIE, compared to the obtained results using the same dataset for training and test, which were 1.81 and 1.82 for the FERET, 2.64 and 2.99 for the Multi-PIE. Indeed, the reason of this outcome is related to the difference between the data distribution of the Multi-PIE and FERET datasets. Ear images from the Multi-PIE and FERET datasets have the same context information, since surrounding area of the ear, e.g., hair, head, have similar structure in almost all datasets. However, the ground truth images that are employed for the comparison with reconstructed images have characteristic features inherent in the corresponding dataset. Because of that, during face reconstruction in the cross-dataset experiments, generative model tends to reconstruct face images with similar background and colors as the ones in the dataset used for training. This inference causes relatively high style difference between reconstructed and ground truth images. 
In addition to this, for test on the Multi-PIE dataset using the FERET model, the obtained performance is better than vice versa experiment and the evaluation metrics are almost the same with the original results.

\section{Conclusion}
In this work, we presented a novel study on biometric modality mapping, i.e. explored a mapping to reconstruct a frontal face image from an ear image input. We formulated the problem as a paired image-to-image translation task and investigated the learning capability of the GAN for deep biometric modality mapping. We employed style reconstruction and face reconstruction losses, in addition to adversarial and pixel losses. We tested our model on the Multi-PIE and FERET datasets using both subject dependent and subject independent experimental setups. We also performed cross-dataset experiments to analyze the generalization capability of the proposed method. Moreover, we conducted face recognition experiments using reconstructed face images and original face images to assess the usability of the generated faces for person identification purposes. According to the experimental results, although there are some artifacts in some local parts of the generated faces, they are still very similar to the original face images. This outcome is also quantitatively measured using five different metrics. These results indicate that the GAN model can learn indirect relationship between ear and face modalities. Using a large-scale dataset that contains high appearance variations would increase the quality of the reconstructed face images and generalization of the model as well. Besides, face recognition performance is found to be very promising, especially on the FERET dataset, on which we have a better reconstruction performance. This outcome also validates the obtained face reconstruction quality. In our future study, to improve the reconstruction quality, remove the local artifacts, and enhance the generalization capacity of the model, we will work further on the proposed network. We also plan to collect a large-scale ear-face image pairs dataset in the wild to capture more appearance variations.


\bibliographystyle{unsrt}  
\bibliography{egbib2}  




\end{document}